\documentclass[journal]{IEEEtran}
\usepackage{amsmath,amsfonts}
\usepackage{algorithmic}
\usepackage{algorithm}
\usepackage{array}
\usepackage[caption=false,font=normalsize,labelfont=sf,textfont=sf]{subfig}
\usepackage{textcomp}
\usepackage{stfloats}
\usepackage{url}
\usepackage{verbatim}
\usepackage{graphicx}
\usepackage{cite}
\usepackage{booktabs}
\usepackage{multirow}
\usepackage[normalem]{ulem}
\usepackage[table]{xcolor}
\begin{document}

\title{UAV-MapFusion: RTK-Aligned Uncertainty-Aware Coarse-to-Fine Multi-Session UAV Mapping}

\author{Feng Pan$^{1*}$, Chunran Zheng$^{2*}$, Bing Xue$^{1}$, Yukang Cui$^{3}$, Jiayu Wen$^{3}$, Zhiyu Chen$^{3}$, and Wei Wang$^{1\dagger}$
\thanks{*These authors contributed equally to this work.}%
\thanks{$^{\dagger}$Corresponding author.}%
\thanks{$^{1}$Feng Pan, Bing Xue, and Wei Wang are with the College of Automation, Harbin Engineering University, Harbin 150001, China. (e-mail: fengpan97618@163.com; xuebing@hrbeu.edu.cn; wangwei407@hrbeu.edu.cn)}%
\thanks{$^{2}$Chunran Zheng is with Department of Mechanical Engineering, University of Hong Kong, Hong Kong SAR, China. (e-mail: zhengcr@connect.hku.hk)}%
\thanks{$^{3}$Yukang Cui, Jiayu Wen, and Zhiyu Chen are with College of Mechatronics and Control Engineering, Shenzhen University, Shenzhen 518060, China. (e-mail: cuiyukang@gmail.com)}
\thanks{This paper has been accepted by IEEE Robotics and Automation Letters.}
\thanks{\copyright{} 2026 IEEE. Personal use of this material is permitted. For all other uses, permission from IEEE is required.}
}


\IEEEpubid{}

\maketitle

\begin{abstract}
Large-scale point cloud maps are essential for robotics and spatial intelligence tasks. UAVs provide an efficient means for large-scale map acquisition; however, due to limited flight endurance and onboard storage, mapping a large-scale scene within a single flight remains difficult. Existing multi-session map merging methods can extend the mapping range, yet in UAV scenarios they still struggle to simultaneously suppress long-range drift and preserve local geometric accuracy. To address this issue, an uncertainty-aware multi-session point cloud map merging and coarse-to-fine optimization system is proposed. The proposed method first performs initial multi-session map merging based on a scene graph, and then incorporates RTK observations through an RTK spatiotemporal alignment module, where temporal offsets are estimated using Dynamic Time Warping (DTW), and continuous RTK constraints are recovered using Multi-Output Gaussian Processes (MOGP) under incomplete sampling and frame dropouts. On this basis, a unified uncertainty-aware factor graph is constructed, and local geometric accuracy is further improved through iterative plane-factor refinement. Experiments on real-world datasets validate the effectiveness and robustness of the proposed method. To facilitate further research and development in the community, our code and dataset will be publicly released.
\end{abstract}


\section{Introduction}
\IEEEPARstart{L}{arge-scale} point cloud maps are fundamental to the deployment of robotics, spatial intelligence, and digital twin applications in large-scale environments, as they provide high-precision 3D geometric representations and physical constraints for real-world digital modeling and intelligent perception \cite{cadena2017past, biljecki2015applications, whitty2010autonomous, zheng2026fast}. Compared with ground-based platforms, unmanned aerial vehicles (UAVs), benefiting from their high mobility and ease of deployment, have become an important platform for acquiring large-scale point cloud maps \cite{ren2025survey}. However, due to limitations in flight endurance, onboard storage, and payload capacity, a single UAV is generally unable to complete comprehensive mapping of a large-scale scene within a single flight \cite{wei2024large, hu2024ms}.

Due to the limited spatial coverage of a single flight, large-scale scene mapping usually requires multiple UAV flights to achieve sufficient coverage. This naturally leads to a multi-session mapping problem, where maps acquired in different flights need to be aligned and merged into a unified map. By jointly aligning and optimizing these session maps, multi-session map merging enables large-scale scene coverage and increases the flexibility of data collection. However, it remains challenging in UAV scenarios. Trajectories within individual sessions are often affected by accumulated drift; inter-session overlap can become sparse or ambiguous due to viewpoint changes and temporal variations; meanwhile, the reliability of heterogeneous measurements may vary dynamically with flight conditions and scene characteristics. As a result, global alignment and local fine registration are difficult to achieve in a unified manner.

Existing studies on multi-session map merging can be broadly divided into two categories. The first category comprises place recognition methods, such as Scan Context++ \cite{kim2021scan}, BTC \cite{yuan2024btc}, iBTC \cite{zou2024ibtc}, RING++ \cite{xu2023ring++} and PointNetVLAD \cite{uy2018pointnetvlad}, which improve the robustness of inter-session place recognition and coarse pose estimation mainly through structured representations or global feature learning. However, these methods typically provide only correspondences and lack a complete map merging mechanism. The second category includes system-level multi-session map fusion frameworks, such as LAMM \cite {wei2024large} and MS-Mapping \cite{hu2024ms}. These methods integrate key components such as data association, outlier rejection, and global optimization, but are mostly designed for ground platforms, such as wheeled vehicles and handheld devices, and are therefore not well suited to the severe viewpoint changes caused by 6-DoF UAV motion. Moreover, they rely primarily on LiDAR measurements and lack heterogeneous constraints, which limits their ability to suppress long-term drift in large-scale environments. Although systems such as Osprey \cite{Osprey} support repeated UAV flights for mapping, they place greater emphasis on integrating existing modules and have not been specifically designed for large-scale UAV point cloud map merging. Therefore, in UAV-based large-scale mapping scenarios, a systematic solution that can jointly ensure reliable inter-session association, globally consistent optimization, and locally accurate registration is still lacking.

To address the aforementioned challenges, we propose an uncertainty-aware multi-session point cloud map merging and coarse-to-fine optimization system for large-scale UAV mapping. Specifically, we incorporate RTK information into the multi-session merging framework and design a spatiotemporal alignment mechanism to mitigate large-scale drift caused by long-term operation, addressing issues such as time offset and frame loss. Furthermore, we establish a unified uncertainty-aware optimization framework, enabling multi-source constraints to be adaptively weighted based on observation quality, thereby improving the robustness of joint optimization. Finally, we design an effective plane factor to enable fine-grained local refinement after coarse global alignment. The main contributions of this paper are summarized below:
\begin{itemize}
    \item We propose an uncertainty-aware multi-session point cloud map merging and coarse-to-fine optimization system for UAV-based large-scale mapping, which enables robust global alignment and locally accurate registration of cross-session maps within a unified framework.

    \item We incorporate RTK information into the multi-session map merging process and design a spatiotemporal alignment mechanism to handle temporal offsets and frame dropouts, thereby alleviating large-scale drift.

    \item We establish a unified uncertainty-aware multi-source factor graph optimization framework, in which adaptive weighting is introduced to improve the robustness of joint optimization. In addition, a plane constraint factor is designed to further enable local geometric refinement and improve local map accuracy.

\end{itemize}
\section{Related Work}
\subsection{Inter-Session Association}

Reliable inter-session association is a prerequisite for multi-session map merging, since the quality of subsequent optimization is determined by the reliability of cross-session correspondences. Early studies mainly relied on handcrafted LiDAR descriptors. Scan Context enabled efficient large-scale retrieval under yaw changes~\cite{kim2021scan}, while BTC further improved robustness by combining local binary cues with triangle-based geometric relations~\cite{yuan2024btc}. Later, learning-based methods such as PointNetVLAD and MinkLoc3D improved descriptor discriminability by learning global point-cloud representations~\cite{uy2018pointnetvlad, MinkLoc3D}, and RING++ extended this line toward global localization under large viewpoint changes~\cite{xu2023ring++}. However, these methods mainly provide candidate matches or coarse pose cues, rather than stable inter-session constraints for multi-session fusion. To improve robustness beyond retrieval, geometric verification and consistency reasoning were further introduced. \cite{Pairwise} selected mutually consistent inter-map measurements to suppress false associations, and FRAME combined overlap detection with rigid registration refinement for autonomous map merging~\cite{FRAME}. These studies suggest that reliable association requires not only place recognition, but also geometric validation and outlier rejection. 

Still, most existing methods remain focused on pairwise matching or registration. As a result, they cannot reliably establish globally consistent inter-session constraints for stable fusion and optimization across multiple sessions.
\subsection{Multi-Session Optimization}

Based on reliable inter-session constraints, a number of multi-session mapping systems have been developed to improve global consistency across repeated runs or multiple map segments. LTA-OM exploited long-term association across sessions~\cite{LTA-OM}, while AutoMerge addressed large-scale map assembly through incremental merging and pose-graph smoothing~\cite{AutoMerge}. More dedicated multi-session LiDAR systems, such as LAMM and MS-Mapping, further combined cross-session loop detection with global optimization to improve large-scale map merging accuracy~\cite{wei2024large,hu2024ms}. \cite{yu2023multi} explored a lightweight multi-session mapping paradigm using semantic line and plane primitives, with a stronger emphasis on localization-oriented map representation than on globally optimized large-scale point-cloud fusion. In related multi-robot settings, LAMP~2.0 also highlighted the importance of a robust back-end for maintaining global consistency in challenging environments~\cite{LAMP2.0}. Although repeated UAV flights were considered in Osprey~\cite{Osprey}, its primary focus was on low-altitude autonomous aerial mapping and next-best-view planning, rather than kilometer-scale multi-session map fusion.

Despite this progress, existing optimization frameworks are still mainly driven by LiDAR constraints. This limitation becomes more pronounced in our UAV setting, where flights are conducted at roughly 80--130\,m altitude over kilometer-scale scenes, causing stronger viewpoint changes and more severe long-range drift across sessions. In addition, RTK observations are available and can provide valuable global-position information, but they cannot be directly fused in practice due to time offset, incomplete sampling, and frame dropouts. Therefore, beyond multi-session LiDAR optimization itself, a practical UAV system should also address RTK spatiotemporal alignment and reliable uncertainty modeling. This is exactly the gap addressed in our work, where RTK observations are aligned and incorporated into an uncertainty-aware coarse-to-fine optimization framework for globally consistent and locally accurate map merging.
\section{Methodology}

\subsection{System Overview}

\begin{figure*}[tbp]
  \centering
  \includegraphics[width=\linewidth]{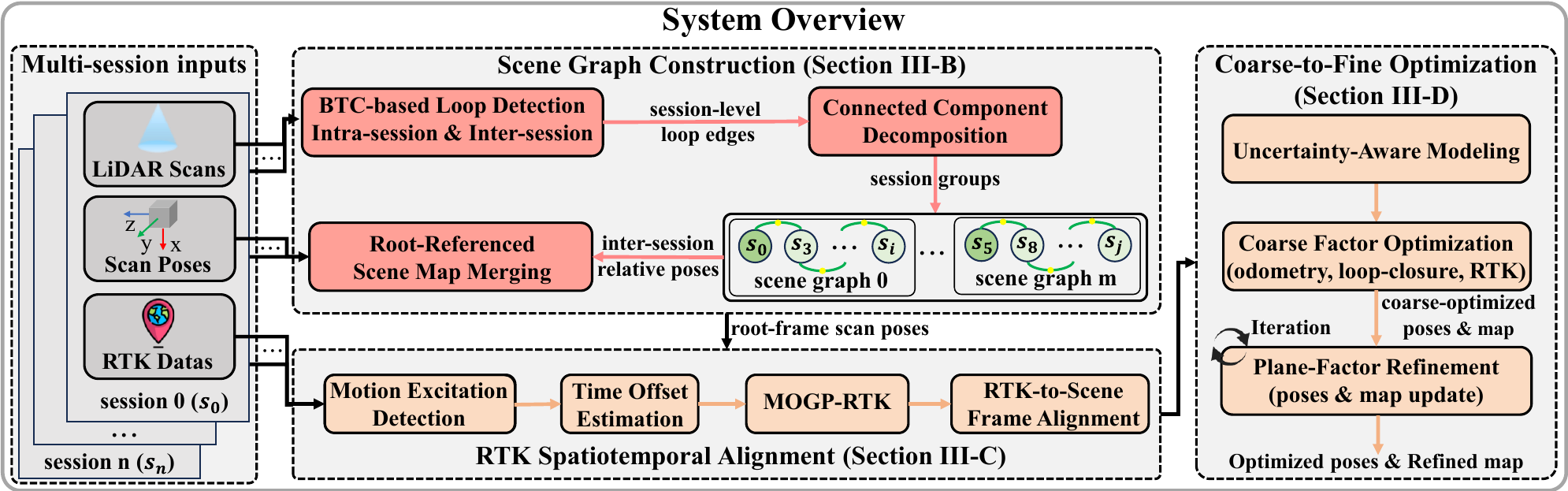}
  \vspace{-0.7cm}
  \caption{System overview of UAV-MapFusion. The input includes multi-session LiDAR scans, RTK measurements, and initial poses from a front-end SLAM system, e.g., FAST-LIVO1/2~\cite{zheng2022fast, zheng2024fast, zhou2025fast}. UAV-MapFusion constructs BTC-based scene graphs, aligns RTK measurements via MOGP-RTK, and performs uncertainty-aware coarse-to-fine optimization. Colors distinguish LiDAR-based scene-graph construction, RTK alignment, and pose-map optimization modules. Black arrows with ``...'' denote session-wise inputs, and circular arrows indicate iterative pose--map updates.}
  \vspace{-0.5cm}
  \label{system_overview}
\end{figure*}

As illustrated in Fig.~\ref{system_overview}, a coarse-to-fine multi-session mapping framework is proposed for large-scale UAV datasets. A set of sessions is taken as input, with each session providing (i) odometry poses from a front-end SLAM system, (ii) RTK measurements in ENU coordinates, and (iii) LiDAR scans. As output, multiple globally consistent scene maps are produced together with optimized poses.

The proposed framework consists of three major components: scene-graph construction, RTK spatiotemporal alignment, and coarse-to-fine factor-graph optimization. First, BTC \cite{yuan2024btc} is employed to establish intra- and inter-session connectivity, based on which connected components are identified to build a scene graph; sessions belonging to the same scene graph are then merged into a unified map. Next, in RTK spatiotemporal alignment, motion-excited segments are extracted and used by DTW to estimate the time offset between RTK and odometry. After temporal compensation, a MOGP is used to predict RTK positions at odometry timestamps to construct synchronized pose–RTK pairs. These pairs enable the estimation of the rigid-body transformation between RTK and the scene coordinate frame, after which all RTK measurements are aligned to the scene frame. Finally, a coarse-to-fine optimization is performed on an uncertainty-aware factor graph: in the coarse stage, odometry, loop-closure, and RTK factors are jointly optimized to correct global drift; in the fine stage, local geometric accuracy is further improved via an iterative plane-factor refinement scheme that alternates between plane reconstruction and pose optimization.

\subsection{Scene Graph Construction}

The constructed scene graph explicitly encodes both intra-session and inter-session loop closures, thereby providing the structural basis for scene-level grouping and subsequent cross-session alignment. 
Specifically, intra-session loop detection is first performed within each session to suppress drift accumulation and to provide reliable anchors for later inter-session association. The LiDAR scan sequence is traversed incrementally, and a new keyframe is selected once the relative translation or rotation increment with respect to the last keyframe exceeds predefined thresholds. For each keyframe, a local submap is constructed by aggregating several neighboring scans before and after the keyframe, and a BTC descriptor is then extracted for place retrieval. The retrieved candidate loop closures are subsequently verified by point-to-plane ICP, and the verified relative poses are retained as reliable intra-session loop constraints. Based on the same keyframes and descriptors, inter-session loop detection is further conducted across different sessions to establish cross-session correspondences. After that, all sessions are abstracted as the node set \(V\) of an undirected graph \(G=(V,E)\), where an edge \((i,j)\in E\) is added if there exists at least one geometrically verified keyframe loop closure between sessions \(s_i\) and \(s_j\). Breadth-first search is then applied to extract the connected component
\[
C(v)=\{u\in V \mid u\ \text{is reachable from}\ v\ \text{in}\ G\},
\]
and each connected component is treated as one scene graph. For each scene graph, the session with the smallest ID is selected as the root session \(s_r\). Since multiple valid keyframe-level loop constraints may exist between two connected sessions, RANSAC is employed to select the geometrically most consistent session-level relative pose \(T_{j\leftarrow i}\in SE(3)\). The poses of all remaining sessions are then propagated to the root frame along the graph connectivity, i.e.,
\begin{equation}
T_{r\leftarrow i}=T_{r\leftarrow j}T_{j\leftarrow i}
\end{equation}

thereby yielding a coarse root-referenced initialization for all sessions in the same physical scene.
\subsection{RTK Spatiotemporal Alignment}
This subsection presents an RTK spatiotemporal alignment method to address cross-sensor data misalignment in multi-session fusion, caused by the lack of hardware synchronization between RTK and odometry and by incomplete RTK sampling.

\subsubsection{Motion Excitation Detection}
To stably estimate the temporal offset between the RTK measurements and the scan poses, motion-excited initial segments are first extracted from the RTK position sequence \(\{(t_k^r,\mathbf{p}_k^r)\}\) and the scan-pose sequence \(\{(t_k^s,\mathbf{p}_k^s)\}\), respectively. Specifically, the velocity magnitude along each timeline is computed from adjacent position differences, and the first timestamp whose velocity exceeds the threshold \(v_{\mathrm{th}}\) is identified in each sequence as \(t_{\mathrm{start}}^x\). Starting from this timestamp, a time window of length \(T\) is then cropped, and a sequence of 3D displacement magnitudes relative to the initial position is constructed. This process provides reliable inputs for the subsequent robust temporal alignment based on DTW.
\subsubsection{Time Offset Estimation}
Based on the motion-excited segments extracted above, DTW is employed to estimate the temporal offset between the RTK and scan-pose sequences. Let the two displacement-magnitude sequences be denoted as $\mathbf{d}^r=\{d_1^r,\dots,d_m^r\}$ and $\mathbf{d}^s=\{d_1^s,\dots,d_n^s\}$, with corresponding timestamps $\{t_i^r\}$ and $\{t_j^s\}$.

\begin{figure}[h]
  \centering
  \includegraphics[width=\linewidth]{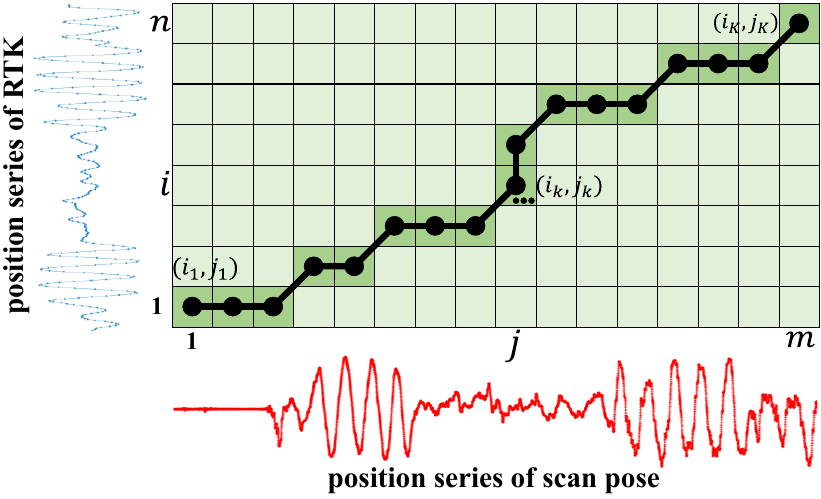}
  \caption{The DTW warping path for temporal offset estimation.}
  \label{fig:dtw_path}
\end{figure}

As illustrated in Fig.~\ref{fig:dtw_path}, DTW searches for an optimal warping path $P=\{(i_k,j_k)\}_{k=1}^{K}$ from the start point $(i_1,j_1)=(1,1)$ to the end point $(i_K,j_K)=(m,n)$ under monotonicity and continuity constraints, by minimizing the cumulative matching cost defined from the local distance $c(i,j)=|d_i^r-d_j^s|$. After obtaining $P$, the timestamp differences of all matched pairs are collected as $\delta_{ij}=t_i^r-t_j^s,\ (i,j)\in P$, and their mean is taken as the temporal offset, i.e., $\Delta t=\frac{1}{|P|}\sum_{(i,j)\in P}\delta_{ij}$. The aligned RTK timestamps are then updated by $t^{r}_{\mathrm{aligned}}=t^{r}+\Delta t$. Within each session, $\Delta t$ is assumed to remain constant.
\subsubsection{MOGP-RTK} 
To overcome RTK asynchrony and frame dropouts, a multi-output Gaussian process is employed to model the RTK trajectory in continuous time, thereby providing temporally consistent and uncertainty-aware position estimates for subsequent coordinate frame alignment.

\textbf{Local training set construction.} Let the time-aligned RTK sequence be $\{(t_i^r,\mathbf{p}_i^r)\}$, where $\mathbf{p}_i^r=[x_i^r,y_i^r,z_i^r]^\top$. For a query time $t^\ast=t_j^s$, only the 10 nearest RTK observations before and after $t^\ast$ on the time axis are selected to form a local training set
\begin{equation}
\mathcal{D}^\ast=\{(t_i^r,\mathbf{p}_i^r)\mid i\in\mathcal{N}(t^\ast)\},
\end{equation}
where $\mathcal{N}(t^\ast)$ denotes the corresponding index set. This local on-demand modeling reduces the cost of fitting a global GP and better accommodates non-uniform RTK sampling.

\textbf{Multi-output modeling with shared hyperparameters.} The three coordinates are modeled independently while sharing the same kernel hyperparameters and global noise scale. For each dimension $a\in\{x,y,z\}$,
\begin{equation}
y_i^a=f_a(t_i^r)+\epsilon_i^a,\qquad
\epsilon_i^a\sim\mathcal{N}(0,\lambda^2\sigma_{a,i}^2),
\end{equation}
where $\lambda$ is the global noise scale. The RTK-reported accuracy is used to define heteroscedastic noise, i.e., $\sigma_{x,i}=\sigma_{y,i}=\sigma_{h,i}$ and $\sigma_{z,i}=\sigma_{v,i}$. An RBF kernel is adopted:
\begin{equation}
k(t,t')=\sigma_f^2\exp\!\left(-\frac{(t-t')^2}{2\ell^2}\right),
\end{equation}
and the shared hyperparameters $\theta=\{\ell,\sigma_f,\lambda\}$ are estimated on $\mathcal{D}^\ast$ by maximizing the log marginal likelihood.

\textbf{Prediction with uncertainty.} For a query time $t^\ast$, the posterior in each dimension is
\begin{equation}
p(f_a(t^\ast)\mid\mathcal{D}^\ast)=\mathcal{N}(\mu_a^\ast,(\sigma_a^\ast)^2),
\end{equation}
with
\begin{equation}
\begin{aligned}
\mu_a^\ast &= \mathbf{k}_\ast^\top(\mathbf{K}+\mathbf{R}_a)^{-1}\mathbf{y}_a, \\
(\sigma_a^\ast)^2 &= k(t^\ast,t^\ast)-\mathbf{k}_\ast^\top(\mathbf{K}+\mathbf{R}_a)^{-1}\mathbf{k}_\ast .
\end{aligned}
\end{equation}
where $\mathbf{K}$ is the kernel matrix built from the local timestamps $\{t_i^r\}$, $\mathbf{k}_\ast$ is the kernel vector between $t^\ast$ and the training points, $\mathbf{y}_a$ is the observation vector of dimension $a$, and $\mathbf{R}_a=\lambda^2\mathrm{diag}(\sigma_{a,1}^2,\ldots,\sigma_{a,N}^2)$. By stacking the three predictive means and variances, the RTK prediction at $t^\ast$ is obtained as
\begin{equation}
\begin{aligned}
\hat{\mathbf{p}}^r(t^\ast)
&=
\begin{bmatrix}
\mu_x^\ast & \mu_y^\ast & \mu_z^\ast
\end{bmatrix}^{\!\top}, \\
\mathbf{\Sigma}^r(t^\ast)
&=
\mathrm{diag}\!\left((\sigma_x^\ast)^2,(\sigma_y^\ast)^2,(\sigma_z^\ast)^2\right).
\end{aligned}
\end{equation}
Therefore, synchronized and uncertainty-aware RTK observations can be provided at arbitrary scan-pose timestamps.
\subsubsection{RTK-to-Scene Frame Alignment}
Based on the above results, the rigid transformation from the RTK (ENU) frame to the scene root frame is then estimated for each session. Specifically, within the start-up motion window, the MOGP-predicted RTK positions $\hat{\mathbf{p}}^{r}(t_j^{s})$ are paired with the corresponding scan-pose translations $\mathbf{p}^{\mathrm{root}}(t_j^{s})$ at scan-pose timestamps $t_j^{s}$ to form a set of temporally aligned 3D correspondences. The session-wise extrinsic parameters $(\mathbf{R}_s,\mathbf{t}_s)$ are then solved in closed form using SVD. Subsequently, all RTK predictions together with their associated uncertainties are consistently transformed into the root frame, thereby providing accurate inputs for the construction of subsequent RTK factors.

\begin{figure}[t]
  \centering
  \includegraphics[width=\linewidth]{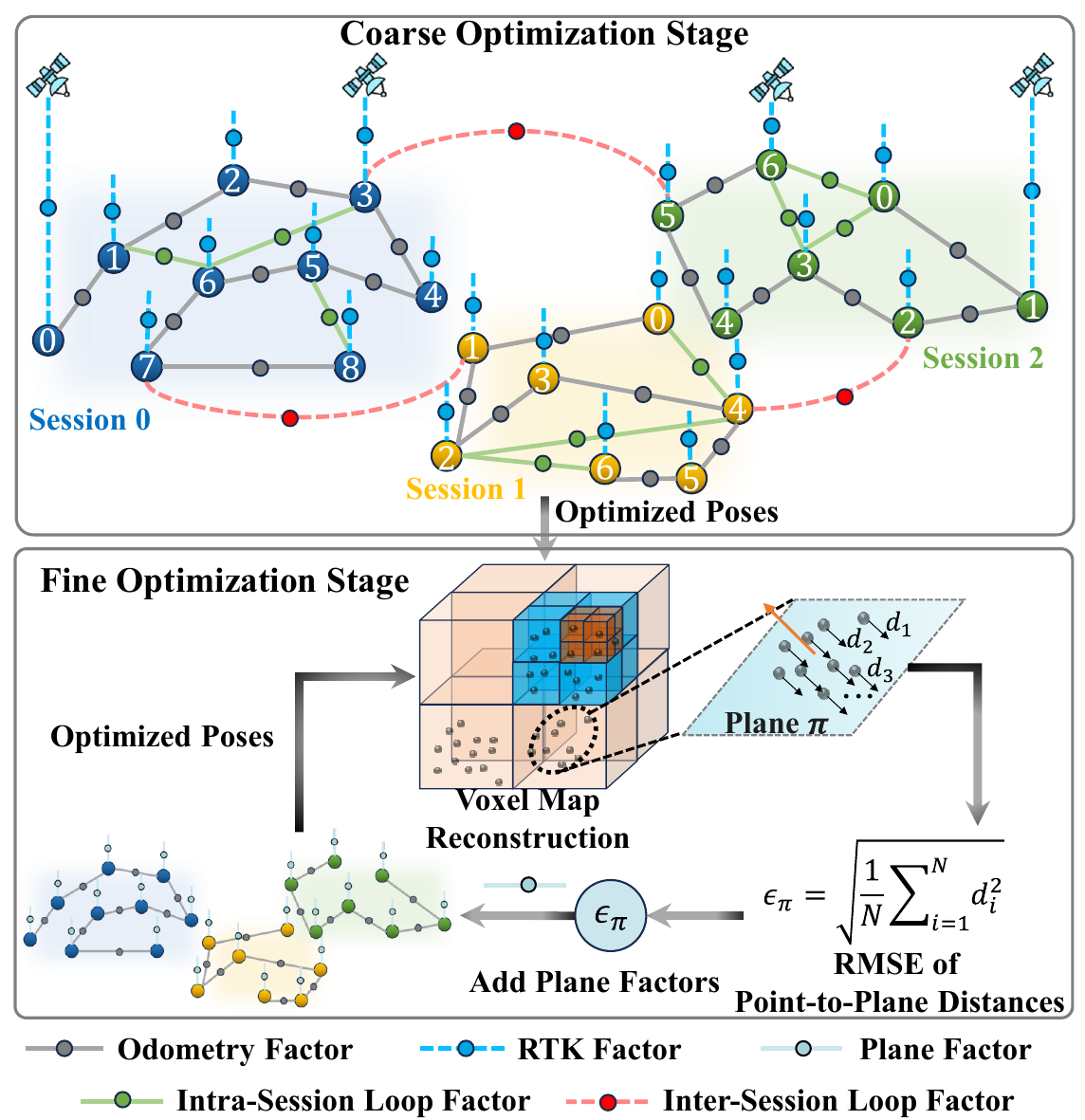}
    \vspace{-0.2cm}
  \caption{Coarse-to-fine optimization workflow. The coarse stage performs uncertainty-aware global optimization with odometry, loop-closure, and RTK factors to correct large-scale drift. The fine stage reconstructs a voxel map and iteratively refines the poses with plane factors derived from point-to-plane RMSE, improving local geometric accuracy and structural sharpness.}
  \label{Coarse to Fine Opti}
    \vspace{-0.3cm}
\end{figure}
\subsection{Coarse-to-Fine Optimization}
Building upon the above modules, an uncertainty-aware coarse-to-fine optimization is constructed to first suppress large-scale drift and then further improve local geometric accuracy, as shown in Fig.~\ref{Coarse to Fine Opti}.

\subsubsection{Uncertainty-Aware Modeling}
An uncertainty-aware weighting scheme is adopted for all factors. Following MS-Mapping, the uncertainty of odometry and loop-closure constraints is estimated from the corresponding registration results and used for adaptive weighting. For RTK factors, the predictive covariance $\mathbf{\Sigma}^{root}(t_j^s)$ produced by MOGP is directly used as the factor noise, such that both the original GNSS measurement quality and the temporal-regression uncertainty are propagated into optimization. For plane factors, the reliability of the $k$-th frame is characterized by plane thickness: for a plane cluster $\pi$, its thickness is defined as $h(\pi)=\sqrt{\lambda_3(\pi)}$, where $\lambda_3(\pi)$ is the smallest eigenvalue of the covariance of the cluster points; accordingly, the frame-level noise is defined as
\begin{equation}
\sigma_k=
\sqrt{
\frac{1}{|P_k|}
\sum_{p\in P_k} h^2(\pi(p))
},
\end{equation}
where $P_k$ denotes the set of planar points in frame $k$. In this way, heterogeneous constraints are weighted according to their observation quality within a unified framework.

\subsubsection{Coarse Factor Graph Optimization}
After scene-graph partition and root-frame alignment, all sessions within the same scene graph are represented in a common coordinate frame. For each scene graph, a coarse factor graph is constructed by jointly incorporating odometry factors, intra-session loop-closure factors, inter-session loop-closure factors, and RTK position factors, with their noises given by the above uncertainty models. The resulting graph is then optimized using GTSAM \cite{dellaert2012factor}, thereby correcting large-scale accumulated drift and providing a reliable initialization for the subsequent fine optimization stage.

\subsubsection{Iterative Plane-Factor Refinement}
Although the coarse optimization removes the dominant global drift, local misalignment may still persist, typically appearing as plane thickening and blurred structural boundaries. To further improve local map quality, an iterative plane-factor refinement is performed on top of the FAST-LIVO2 \cite{zheng2024fast} map structure. Specifically, the global map is organized using the original hierarchical voxel representation, from which local planar clusters are extracted, and neighboring clusters belonging to the same physical plane are further merged by region growing to obtain more complete and stable plane primitives.

For the $k$-th frame, let $\{d_{k,m}\}_{m=1}^{M_k}$ denote the point-to-plane distances from its planar points to the associated merged plane clusters. A frame-level plane factor is then defined by the RMSE residual
\begin{equation}
r_k^{plane}=
\sqrt{
\frac{1}{M_k}
\sum_{m=1}^{M_k} d_{k,m}^2
}.
\end{equation}
Together with odometry factors between consecutive frames, these plane factors are iteratively optimized in GTSAM with the first pose of the root session fixed. After each optimization round, the global map, planar clusters, and point-to-plane associations are rebuilt using the updated poses, and the planarity criterion is progressively tightened until convergence. In this manner, local plane thickness is continuously reduced and structural sharpness is further improved.

\section{Experiments}

{In this section, the proposed method is evaluated on both private datasets and the public MARS-LVIG~\cite{li2024mars} dataset to verify its effectiveness and mapping accuracy. All experiments are performed on a desktop computer with an Intel i9-13900K CPU and 64 GB RAM. Additional implementation details and parameter settings are provided in Table~S1 in the Supplementary Material~\cite{uavmapfusion_supp}.}

\begin{figure}[!h]
  \centering
  \includegraphics[width=\linewidth]{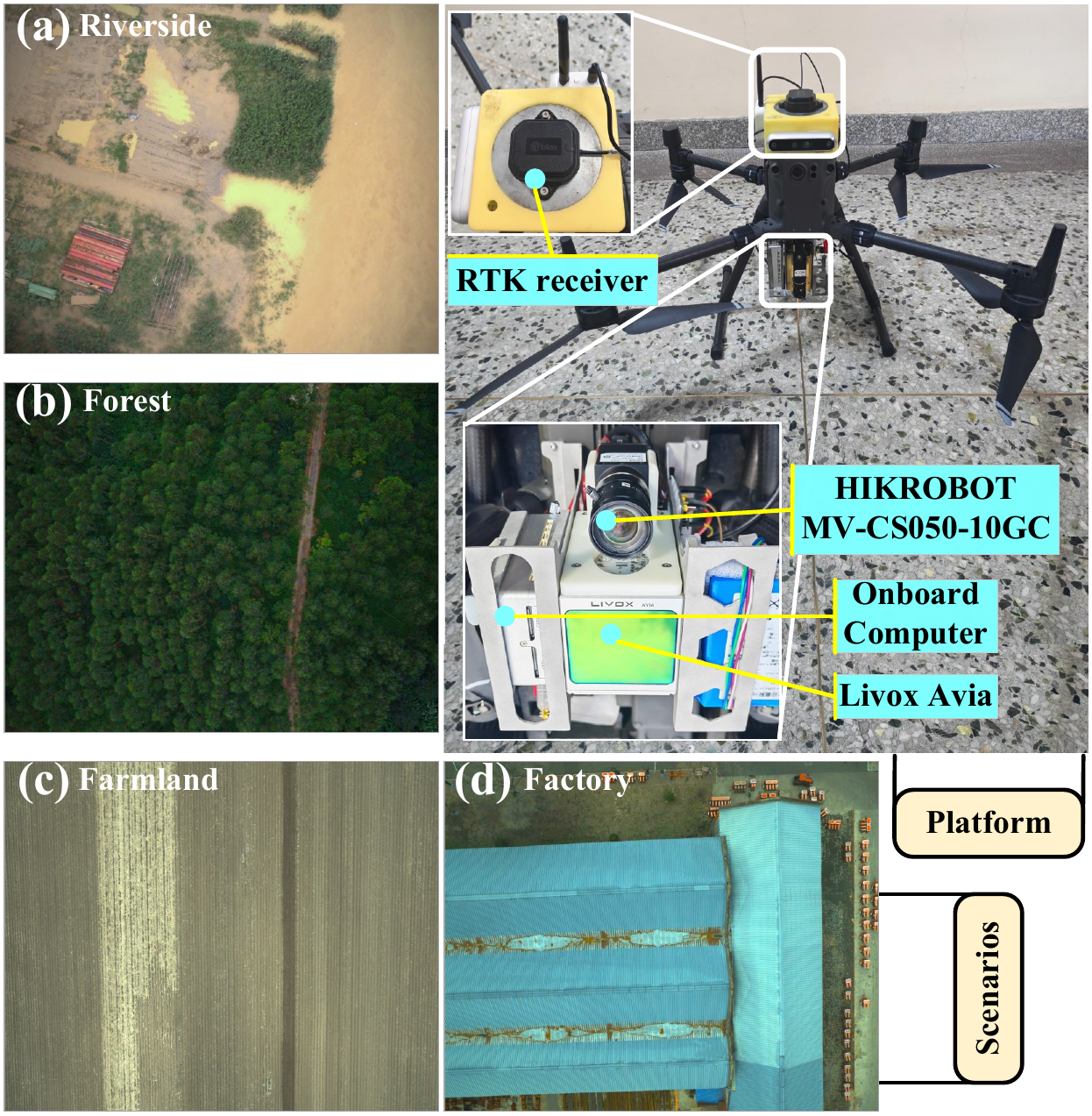}
  \vspace{-0.5cm}
  \caption{Self-developed data acquisition platform. The complete sensing system was independently designed and built in-house. (a) and (b) were collected in Shenzhen, (c) and (d) in Jiamusi, Heilongjiang. All datasets, hardware designs, and deployment details will be open-sourced.}
  \vspace{-0.5cm}
  \label{fig:platform}
\end{figure}
\subsection{Datasets and Setup}
The private datasets are largely consistent with the public MARS-LVIG \cite{li2024mars} dataset in terms of platform configuration and acquisition scale. Both are collected by a DJI M300 RTK UAV equipped with a Livox Avia LiDAR, a Hikvision camera, and a u-blox ZED-F9P receiver. The flight altitude is approximately 80--130 m, and the coverage of each scene ranges from $9.4\times10^4$ to $5.77\times10^5$ m$^2$. In total, eight scenes are included in the evaluation, with four from our self-collected dataset and four from MARS-LVIG. Each scene is composed of four partially overlapping sessions. In addition, the scan poses and LiDAR colorized scans used in the experiments are generated by FAST-LIVO2.

\subsubsection{Private Dataset}
The private dataset includes four representative scenes: Riverside (\textit{Scene1}), Forest (\textit{Scene2}), Factory (\textit{Scene3}), and Farmland (\textit{Scene4}). These scenes represent typical aerial surveying environments, including open riverside areas with scattered shrubs, forested regions with dense vegetation and repetitive geometric structures, industrial areas with abundant man-made structures, and farmland areas with weak texture and large planar surfaces. The acquisition platform of the private dataset is shown in Fig.~\ref{fig:platform}.

\subsubsection{Public Dataset}
The MARS-LVIG dataset contains four scenes: Aero-model airfield (\textit{Scene5}), Island (\textit{Scene6}), Rural town (\textit{Scene7}), and Valley (\textit{Scene8}). These scenes exhibit diverse aerial surveying characteristics, including highly open hard-surface areas, island environments with over-sea flight trajectories and stronger attitude disturbances, rural settlement areas with mixed natural and man-made structures, and mountainous regions with significant terrain variations. For each scene, the sequence with the largest spatial coverage is selected for evaluation in this paper.

\subsection{Evaluation Metrics}
{Following MapEval [22], the Average Wasserstein Distance (AWD) and the Spatial Consistency Score (SCS) are adopted to evaluate the reconstructed maps in terms of global and local consistency, respectively. In our experiments, the reference maps are independently generated from DJI Zenmuse L1 data\footnote{\url{https://www.dji.com/sg/support/product/zenmuse-l1}.} using DJI Terra\footnote{\url{https://enterprise.dji.com/dji-terra}.}, following its standard LiDAR point-cloud reconstruction workflow. No result-dependent manual alignment is applied, and all methods follow the same preprocessing protocol before computing AWD and SCS. To complement the map-level evaluation, we further provide trajectory-level ATE/RPE ablation results, ATE comparisons with LAMM and MS-Mapping, and module-wise runtime and memory-usage analyses in the Supplementary Material~\cite{uavmapfusion_supp}.}

Specifically, for each occupied voxel $v_i$, the point distributions in the ground-truth map and the reconstructed map are modeled as Gaussian distributions, denoted by $\mathcal{N}_i^g(\boldsymbol{\mu}_i^g,\mathbf{\Sigma}_i^g)$ and $\mathcal{N}_i^e(\boldsymbol{\mu}_i^e,\mathbf{\Sigma}_i^e)$, respectively. Their discrepancy is measured by the voxel-wise Wasserstein distance
\begin{equation}
\begin{aligned}
W_i
&=
\Bigg(
\left\|\boldsymbol{\mu}_i^g-\boldsymbol{\mu}_i^e\right\|_2^2 \\
&\quad+
\mathrm{tr}\!\left(
\mathbf{\Sigma}_i^g+\mathbf{\Sigma}_i^e
-2\left(
(\mathbf{\Sigma}_i^e)^{1/2}\mathbf{\Sigma}_i^g(\mathbf{\Sigma}_i^e)^{1/2}
\right)^{1/2}
\right)
\Bigg)^{1/2}.
\end{aligned}
\end{equation}
Then, AWD is defined as
\begin{equation}
\mathrm{AWD}=\frac{1}{M}\sum_{i=1}^{M}W_i,
\end{equation}
where $M$ is the number of occupied voxels. To further quantify local consistency, SCS is computed as
\begin{equation}
\mathrm{SCS}=\frac{1}{M}\sum_{i=1}^{M}\frac{\sigma(\mathcal{W}_{\mathcal{N}(i)})}{\mu(\mathcal{W}_{\mathcal{N}(i)})},
\end{equation}
where $\mathcal{W}_{\mathcal{N}(i)}$ denotes the set of Wasserstein distances in the neighborhood of voxel $v_i$, and $\sigma(\cdot)$ and $\mu(\cdot)$ denote the standard deviation and mean, respectively. Lower AWD and SCS indicate better global and local consistency, respectively. More details can be found in \cite{MapEval}.
\begin{figure*}[!t]
  \centering
  \includegraphics[width=0.86\linewidth]{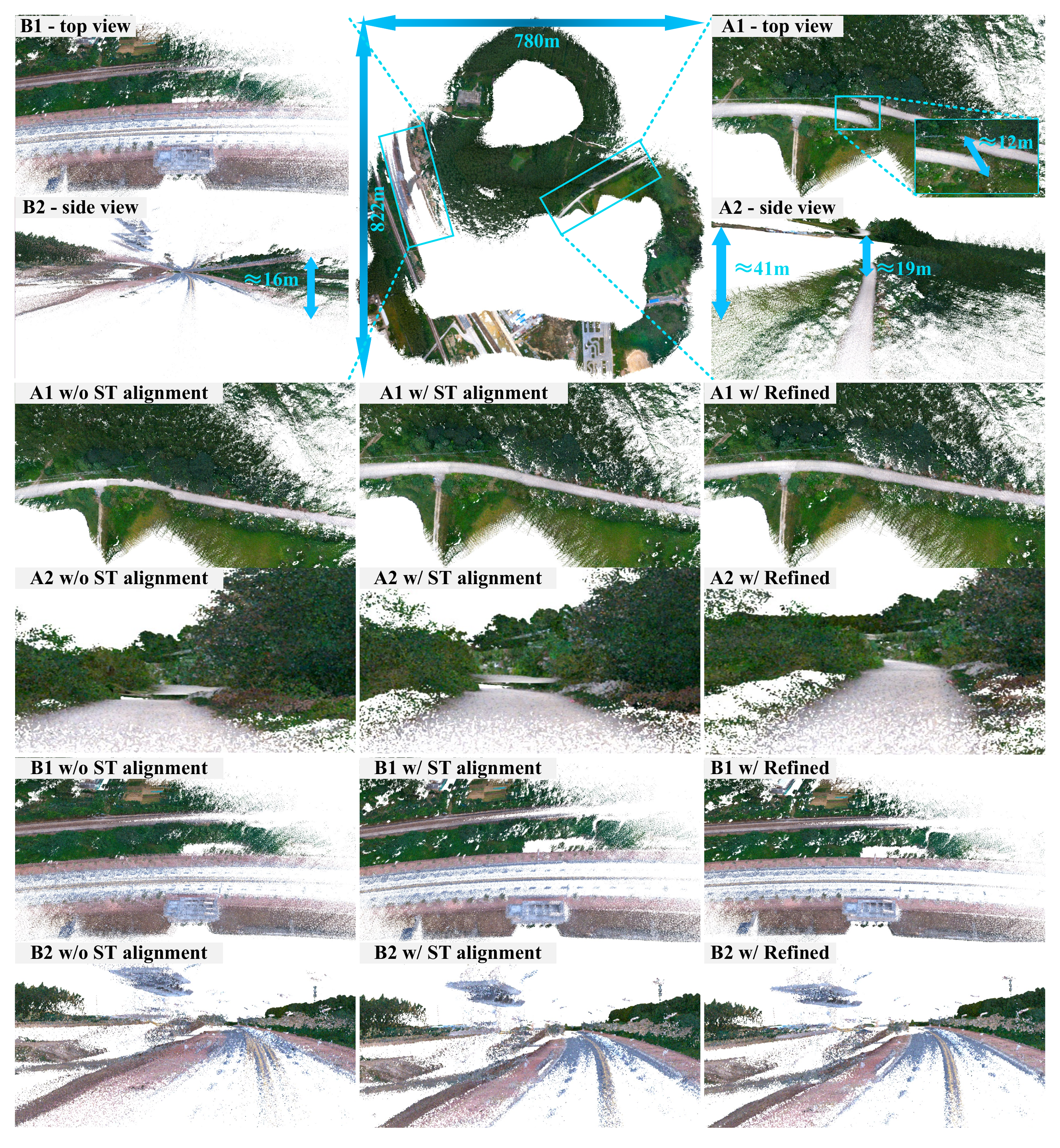}
  \vspace{-0.5cm}
  \caption{Qualitative ablation results on {S2 (Forest)}, a representative forest scene with clear inter-session overlap and severe pre-optimization misalignment. In the overlapping regions, obvious point-cloud displacement and layering can be observed in the raw merged map, with offsets from over ten meters to about 41\,m. RTK spatiotemporal alignment improves global consistency, while fine optimization further refines local structures.}
    \vspace{-0.5cm}
  \label{Ablation Experiment}
\end{figure*}
\subsection{Ablation Experiments}
We conducted ablation experiments on the RTK spatiotemporal alignment in the coarse optimization stage and the plane-factor-based fine optimization, and the results are reported in Table~\ref{tab:ablation_map_eval_vertical}.  Although RTK measurements without spatiotemporal (ST) alignment can still improve global consistency to some extent, the local consistency is clearly degraded. This is because the raw RTK measurements are not strictly synchronized with the odometry timestamps and may also suffer from frame dropouts, such that directly using them introduces biased global constraints and distorts local geometry during optimization. After RTK spatiotemporal alignment is introduced, AWD is further reduced in all scenes, and SCS is also effectively restored, indicating that reliable RTK constraints improve global consistency while preserving local structural coherence. After further incorporating the plane-factor-based fine optimization, both AWD and SCS are significantly reduced, demonstrating that the fine stage is essential for suppressing residual local misalignment and refining local structures. Qualitative comparisons on S2 {(Forest)} are shown in Fig.~\ref{Ablation Experiment}.
\begin{table}[h]
\centering
\caption{Ablation study on eight scenes.}
\label{tab:ablation_map_eval_vertical}
\setlength{\tabcolsep}{4.5pt}
\renewcommand{\arraystretch}{1.05}
\begin{tabular}{lccccc}
\toprule
Scene & Metric & Raw & Coarse w/o ST & Coarse w/ ST & Coarse+Fine \\
\midrule

\multirow{2}{*}{S1}
 & AWD $\downarrow$ & 0.48732 & 0.24891 & 0.10344 & \cellcolor{green!30} 0.00121 \\[2pt]
 & SCS $\downarrow$ & 0.42156 & 0.47238 & 0.39827 & \cellcolor{yellow!30} 0.09824 \\
\midrule
\multirow{2}{*}{S2}
 & AWD $\downarrow$ & 0.53429 & 0.31221 & 0.14461 & \cellcolor{green!25} 0.00176 \\[2pt]
 & SCS $\downarrow$ & 0.45083 & 0.52171 & 0.46955 & \cellcolor{yellow!25} 0.13433 \\
\midrule
\multirow{2}{*}{S3}
 & AWD $\downarrow$ & 0.60385 & 0.28673 & 0.09718 & \cellcolor{green!40} 0.00084 \\[2pt]
 & SCS $\downarrow$ & 0.50842 & 0.56194 & 0.40263 & \cellcolor{yellow!40} 0.07195 \\
\midrule
\multirow{2}{*}{S4}
 & AWD $\downarrow$ & 0.44167 & 0.19625 & 0.07236 & \cellcolor{green!50} 0.00053 \\[2pt]
 & SCS $\downarrow$ & 0.39218 & 0.34671 & 0.22854 & \cellcolor{yellow!50} 0.04117 \\
\midrule
\multirow{2}{*}{S5}
 & AWD $\downarrow$ & 0.57142 & 0.30158 & 0.11894 & \cellcolor{green!35} 0.00102 \\[2pt]
 & SCS $\downarrow$ & 0.48637 & 0.54816 & 0.43129 & \cellcolor{yellow!35} 0.08302 \\
\midrule
\multirow{2}{*}{S6}
 & AWD $\downarrow$ & 0.62815 & 0.35462 & 0.17631 & \cellcolor{green!15} 0.00264 \\[2pt]
 & SCS $\downarrow$ & 0.55374 & 0.61248 & 0.49716 & \cellcolor{yellow!15} 0.16783 \\
\midrule
\multirow{2}{*}{S7}
 & AWD $\downarrow$ & 0.46358 & 0.21473 & 0.08157 & \cellcolor{green!45} 0.00068 \\[2pt]
 & SCS $\downarrow$ & 0.40192 & 0.37264 & 0.25138 & \cellcolor{yellow!45} 0.05269 \\
\midrule
\multirow{2}{*}{S8}
 & AWD $\downarrow$ & 0.59274 & 0.33841 & 0.15248 & \cellcolor{green!20} 0.00193 \\[2pt]
 & SCS $\downarrow$ & 0.51783 & 0.57426 & 0.44683 & \cellcolor{yellow!20} 0.14257 \\

\bottomrule
\end{tabular}\par\vspace{2mm}
\begin{minipage}{\linewidth}
\footnotesize \emph{*} {S1 (Riverside), S2 (Forest), S3 (Factory), S4 (Farmland), S5 (Aero-model Airfield), S6 (Island), S7 (Rural Town), and S8 (Valley)}. AWD $\downarrow$ and SCS $\downarrow$ indicate lower-is-better. Darker green and yellow shading denote smaller AWD and SCS values, respectively.
\end{minipage}
\end{table}

{Additional voxel-size ablation results are also provided in the Supplementary Material, and all final experiments use a fixed voxel partition size of 3.0 $\times$ 3.0 $\times$ 3.0 m as an accuracy-efficiency trade-off.}

\subsection{Benchmark Experiments}
To further verify the effectiveness of the proposed method, two representative multi-session map fusion systems (LAMM \cite{wei2024large} and MS-Mapping \cite{hu2024ms}) for benchmark comparison, and the results are shown in Table ~\ref{tab:benchmark_map_eval_vertical}. It can be seen that the proposed method achieves the best AWD and SCS in all eight scenarios, with the improvement in AWD being particularly significant. This is mainly because both baseline methods rely solely on LiDAR internal constraints and lack effective correction from external global observations, thus making them more prone to global registration errors in complex scenarios such as S2 {(Forest)}, S5 {(Aero-model Airfield)}, S6 {(Island)}, and S8 {(Valley)}. In contrast, the proposed method provides reliable global positional constraints for coarse optimization through RTK spatiotemporal alignment, thereby significantly improving the global consistency of the map.

Regarding local consistency, although LAMM and MS-Mapping also achieve good results in scenarios with relatively regular structures (such as S3 {(Factory)}, S4 {(Farmland)}, and S7 {(Rural Town)}), their SCS remains higher than that of the proposed method due to the lack of further fine-grained optimization. In contrast, the proposed method introduces further refined plane factor iterative optimization after coarse optimization, effectively improving the compactness and clarity of the local structure. It should also be noted that LAMM generally outperforms MS-Mapping in overall performance, mainly due to its more robust loop closure detection and mismatch removal mechanisms. Overall, the results show that the proposed coarse-to-fine optimization framework can simultaneously achieve more reliable global alignment and better local map quality.
\begin{table}[h]
\centering
\caption{Benchmark comparison on eight scenes.}
\label{tab:benchmark_map_eval_vertical}
\renewcommand{\arraystretch}{1.15}
\resizebox{\columnwidth}{!}{%
\begin{tabular}{lcccccc}
\toprule
\multirow{2}{*}{Scene} & \multicolumn{2}{c}{Proposed} & \multicolumn{2}{c}{LAMM} & \multicolumn{2}{c}{MS-Mapping} \\
\cmidrule(lr){2-3} \cmidrule(lr){4-5} \cmidrule(lr){6-7}
 & AWD $\downarrow$ & SCS $\downarrow$ & AWD $\downarrow$ & SCS $\downarrow$ & AWD $\downarrow$ & SCS $\downarrow$ \\
\midrule
S1 & \cellcolor{green!30} 0.00121 & \cellcolor{yellow!30} 0.09824 & 0.09237 & 0.23864 & 0.11829 & 0.21473 \\
S2 & \cellcolor{green!25} 0.00176 & \cellcolor{yellow!25} 0.13433 & 0.13658 & 0.31842 & 0.17186 & 0.29435 \\
S3 & \cellcolor{green!40} 0.00084 & \cellcolor{yellow!40} 0.07195 & 0.12846 & 0.19271 & 0.16524 & 0.18153 \\
S4 & \cellcolor{green!50} 0.00053 & \cellcolor{yellow!50} 0.04117 & 0.06138 & 0.13256 & 0.07947 & 0.11728 \\
S5 & \cellcolor{green!35} 0.00102 & \cellcolor{yellow!35} 0.08302 & 0.14762 & 0.26639 & 0.18374 & 0.24155 \\
S6 & \cellcolor{green!15} 0.00264 & \cellcolor{yellow!15} 0.16783 & 0.18943 & 0.35568 & 0.23657 & 0.33246 \\
S7 & \cellcolor{green!45} 0.00068 & \cellcolor{yellow!45} 0.05269 & 0.07432 & 0.14952 & 0.09826 & 0.13863 \\
S8 & \cellcolor{green!20} 0.00193 & \cellcolor{yellow!20} 0.14257 & 0.16449 & 0.30172 & 0.20534 & 0.28659 \\
\bottomrule
\end{tabular}%
}
\par\vspace{2mm}
\begin{minipage}{\linewidth}
\footnotesize \emph{*} {S1 (Riverside), S2 (Forest), S3 (Factory), S4 (Farmland), S5 (Aero-model Airfield), S6 (Island), S7 (Rural Town), and S8 (Valley)}. 
\end{minipage}
\vspace{-0.5cm}
\end{table}

\section{Conclusion}

This paper presents an uncertainty-aware coarse-to-fine optimization system for large-scale UAV mapping. Multi-session data are first unified through scene-graph construction and scene map merging. To address the temporal misalignment and frame dropouts between RTK and odometry, an RTK spatiotemporal alignment method is introduced by combining DTW with MOGP, thereby providing continuous and reliable RTK observations for optimization. On this basis, an uncertainty-aware factor graph is constructed by jointly incorporating odometry, loop-closure, and RTK constraints, and the local geometric quality is further improved through iterative plane-factor refinement. Experiments on both private and public datasets demonstrate that the proposed method effectively improves the global consistency and local structural clarity of large-scale UAV mapping. Nevertheless, the current system still has several limitations. The fine optimization stage relies to some extent on planar structures, and its effectiveness may be constrained in scenes with weak planar geometry or highly irregular surfaces. {In addition, the experiments use FAST-LIVO2 as the odometry frontend, and the generality to other frontend odometry systems remains to be further validated. Severe RTK dropouts around critical map-stitching regions may also lead to residual misalignment when the remaining loop and plane constraints are too weak, as illustrated by the S1 (Riverside) failure case in Fig.~S9 of the Supplementary Material~\cite{uavmapfusion_supp}.} Future work will focus on frontend-general validation, more general geometric constraint modeling, and more robust optimization strategies for complex large-scale UAV mapping scenarios.

\bibliographystyle{IEEEtran}
\bibliography{references}

\end{document}